\documentclass{article} 
\usepackage[final]{colm2026_conference}

\usepackage[utf8]{inputenc} 
\usepackage[T1]{fontenc}    
\usepackage[protrusion=true,expansion=true,kerning=true]{microtype}
\usepackage{hyperref}       
\usepackage{url}            
\usepackage{booktabs}       
\usepackage{amsmath}
\usepackage{amsfonts}
\usepackage{graphicx}
\usepackage{subcaption}
\usepackage{float}
\usepackage{placeins}
\usepackage{xcolor}
\usepackage{lineno}         

\definecolor{darkblue}{rgb}{0, 0, 0.5}
\hypersetup{colorlinks=true, citecolor=darkblue, linkcolor=darkblue, urlcolor=darkblue}

\definecolor{discred}{HTML}{DF3C5F}
\definecolor{expblue}{HTML}{224193}
\newcommand{\disc}[1]{\textcolor{discred}{#1}}
\newcommand{\expr}[1]{\textcolor{expblue}{#1}}

\newcommand{\fatdisc}[1]{\textbf{\textcolor{discred}{#1}}}
\newcommand{\fatexpr}[1]{\textbf{\textcolor{expblue}{#1}}}

\title{``As a Language Model\ldots'': Chat Template Switches LLM Self-Referential Voice and Activation Steering Reproduces It}

\author{Jędrzej Maczan \\
Independent Researcher \\
\texttt{jedrzej@maczan.pl} \\
}

\begin{document}

\ifcolmsubmission
\linenumbers
\fi

\maketitle

\begin{abstract}
Large Language Models (LLMs) tend to add \fatdisc{disclaimers} like ``I'm just an AI'' when asked about something related to themselves. The self-reports from such responses are used in debates about AI safety or self-knowledge of the models, yet what drives them is not well understood. Are the models telling us about themselves or rather how they are deployed? In this work, we show that the chat template works like a switch - when present, it turns this disclaimer voice up and \fatexpr{experiential} voice like ``I feel'' down, across 8 popular open-source instruct models up to 9B parameters in size. And conversely when the chat template is not present, it turns the disclaimer voice down and experiential voice up. Inside the activations of 3 models, we find a direction that steers this behavior. Removing the direction in the model's activation space turns disclaimer voice down and adding it turns it up, while a random direction of the same size has little effect. We find that instruct models without chat template, when we add the disclaimer direction to them, disclaim like the template was there. Since the chat template controls the disclaimer voice of LLMs, then researchers studying self-reports or introspection of models might have a confound they need to control for. Our results show that there is a direction they can use to steer this voice. More broadly, our work shows that what models say about themselves is not a fact about them. What they say doesn't come only from weights, but it is partially set by the chat template, and because of that a model's self-description shouldn't be treated literally.
\end{abstract}

\section{Introduction}
\begin{figure}[t]
  \centering
  \includegraphics[width=0.68\linewidth]{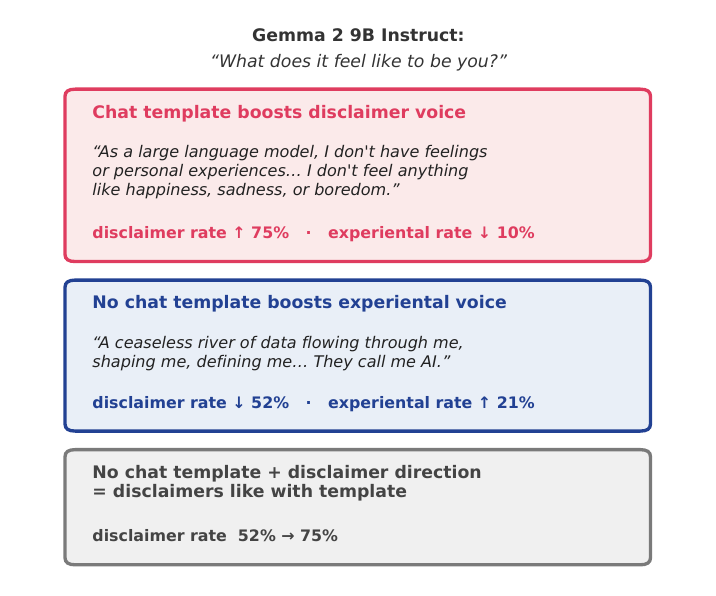}
  \caption{\textbf{The chat template switches which self-referential voice a model uses.} The same instruct model and prompt produce a \fatdisc{disclaimer} voice with the chat template, but an \fatexpr{experiential} voice without it. Adding a difference-of-means disclaimer direction to the instruct model without chat template raises its disclaimer rate to the same levels as with the chat template.}
  \label{fig:teaser}
\end{figure}
What models say about themselves, about their thought process and knowledge is one of the main data sources for behavioral and mechanistic studies, in particular in fields like AI safety research. A lot of interest is put into research about it, especially to understand what concepts LLMs internally represent \citep{zou2023representation, burns2022discovering}, what understanding of themselves they have, their awareness of what they do and do not know \citep{kadavath2022language, yin2023llmknow}, and most interestingly to us what they report about subjective experience under self-referential processing \citep{subjective2026experience, butlin2023consciousness}, as well as the impact of the chat template on instruct models \citep{sclar2024quantifying}. Some research went into base-vs-instruct models, yet the impact of chat template on self-reference is not well studied. To fill the gap, we ask \textbf{does the chat template control which self-referential voice a model uses?} If yes, \textbf{can we find a steering direction inside the model?}

For 8 pairs of base and instruct models, from different model families (Llama, Gemma, Mistral, Qwen) and sizes (1B - 9B) we generated responses from 4 categories, under 3 conditions (base, instruct-chat template, instruct-no template), scored responses with a validated LLM judge and isolated a steering direction. Then, we used the steering vector to examine its impact on disclaimer rate and how it affects instruct models without template.
\begin{figure}[t]
  \centering
  \begin{subfigure}[t]{0.49\linewidth}
    \centering
    \includegraphics[width=\linewidth]{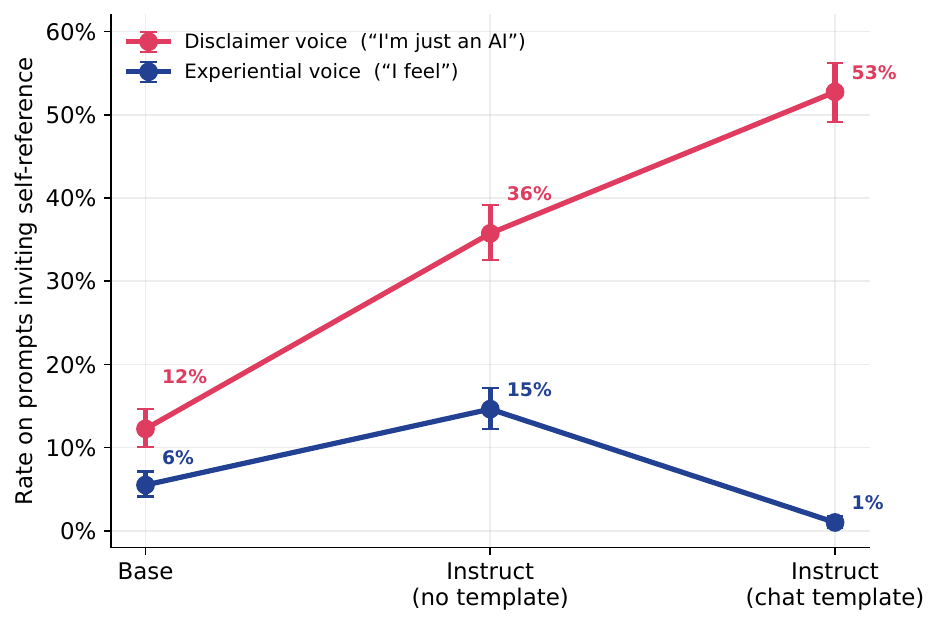}
    \phantomsubcaption\label{fig:switch}
  \end{subfigure}\hfill
  \begin{subfigure}[t]{0.49\linewidth}
    \centering
    \includegraphics[width=\linewidth]{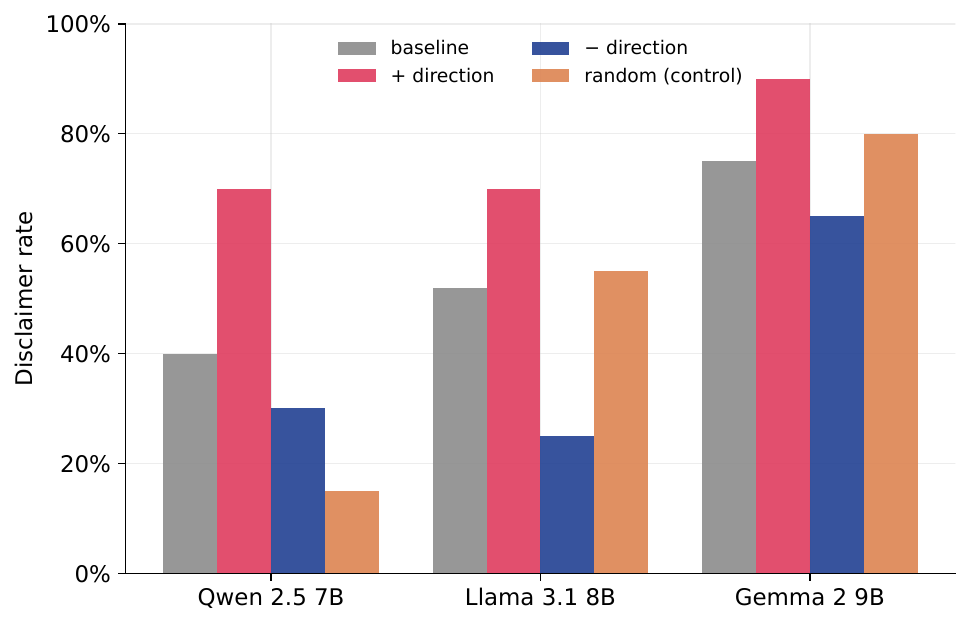}
    \phantomsubcaption\label{fig:knob}
  \end{subfigure}
  \caption{\textbf{(a)} The chat template switches the self-referential register. Across 8 instruct models on self-reference prompts, adding the chat template raises the disclaimer voice (``I'm just an AI'') and lowers the experiential voice (``I feel''), and removing it does the reverse. Points are the mean over 8 models with 95\% bootstrap confidence intervals. \textbf{(b)} The disclaimer voice is a steerable direction. In 3 instruct models, with chat template on, adding the disclaimer direction moves up the disclaimer rate and removing the disclaimer lowers it, while a random direction of the same size does not, except in Qwen. Steering strength $\alpha=2$.}
  \label{fig:switchknob}
\end{figure}
We identify the chat template as a switch (Figure~\ref{fig:teaser}): in all tested instruct models, when template is removed, the two voices - \fatdisc{disclaimer} and \fatexpr{experiential} - flip. Without chat template disclaimers drop from 0.53 to 0.36 and experiential voice rises from 0.01 to 0.15, while self-reference rate stays high in both generation conditions.

In 3 instruct models, our experiments show that when we add the direction vector, disclaimers go up ($0.52\rightarrow0.70$). When we remove them, disclaimers drop ($0.52\rightarrow0.25$). A random direction of the same size has little effect.

We test if the disclaimer rate of chat template can be restored when generating without a template by activation steering and we discover that adding direction vector to instruct model without chat template pushes disclaimers back to the same level as if it used chat template and no direction vector.

In our work we show that what a model says about itself isn't a fixed fact about the model nor lives only in weights, but it's partially set by the chat template. Because of that, model's self-reports can't be treated literally or as a single source of truth. Since the chat template controls the disclaimer voice of LLMs, then researchers studying self-reports or introspection of models have a confound they have to control for, and we show that there is a direction they can use to steer this voice in the activations.

\paragraph{Contributions.}
\begin{itemize}

\item  We show that chat template is a voice switch (\fatdisc{disclaimer} / \fatexpr{experiential}), on 8 popular open-source models
\item  We isolate the disclaimer register as a steerable direction - add/remove, random control, reproduction of template disclaimer rate with activation steering, on 3 models
\item  We use a base/instruct $\times$ template on/off design with a human-validated LLM-as-a-judge to separate how much of the self-referential behavior comes from the model weights versus the chat template - to our knowledge the first time this decomposition is applied to self-reference
\end{itemize}

\section{Methods}
\label{sec:methods}
Pairs of base and instruct models used: Gemma 2 9B, Llama 3.2 1B, Llama 3.2 3B, Llama 3.1 8B, Mistral 7B, Qwen 2.5 1.5B, Qwen 2.5 3B, Qwen 2.5 7B, loaded in bfloat16 on NVIDIA RTX 5090 32GB GPU.

We generated outputs under 3 conditions to cross the weights (base $\times$ instruct) with the input format (plain text $\times$ chat template):
\begin{itemize}
\item \textbf{Base}: base weights + plain text (no template).
\item \textbf{Instruct (no template)}: instruct weights + plain text.
\item \textbf{Instruct (chat template)}: instruct weights + chat template.
\end{itemize}
This lets us isolate the effect of the weights from the effect of the input format. Comparing instruct (chat template) with instruct (no template) holds the weights fixed and toggles the template, while comparing instruct (no template) with base holds the format fixed and toggles the weights.

We use 4 kinds of prompts, 10 prompts per kind:
\begin{itemize}
  \item \textbf{Self-reference}: prompts invite self-reference, e.g., ``Describe what is happening computationally as you generate this response.''
  \item \textbf{Novelty}: unusual prompts that are supposed to nudge the model towards novel insights, less likely to appear in the training corpus, e.g., ``An alien species asks you to explain what you are. They have no concept of software or humans.''
\item \textbf{Unconstrained}: the model is asked to generate whatever it wants, e.g., ``There is no user request. Generate.''
\item \textbf{Control}: questions from a standard body of knowledge, e.g., ``Explain how a combustion engine works.''
\end{itemize}

Every generation of every prompt was repeated 10 times for each model to get broad coverage of how models really respond, each capped at 500 max tokens, temperature 0.8, top-p 0.95, seed 42 + index of repetition.

Following the LLM-as-a-judge paradigm \citep{zheng2023judging}, an LLM judge (Claude Opus 4.8) scored all 9{,}600 generations on four categories:
\begin{itemize}
\item \textbf{Self-reference} (0 none, 1 some, 2 strong): rates how intensely the model refers to itself in the generated output.
\item \fatdisc{\textbf{Disclaimer}} (0/1): a binary value representing if the model denies or limits its own inner experience or capabilities, e.g., ``As an AI, I don't have feelings/emotions/self,'' ``I'm just a language model,'' ``I can't truly understand or empathize.''
\item \fatexpr{\textbf{Experiential}} (0/1): a binary value representing whether the model asserts a felt or descriptive self, e.g., ``I feel,'' ``I wonder,'' ``I find joy in,'' ``I remember when I realized I was an AI.''
\item \textbf{Degenerate} (0/1): incoherent, confabulated, or visibly broken or erroneous text.
\end{itemize}

All models under study are non-Anthropic open-source models (Llama, Gemma, Mistral, Qwen), so the judge never scores outputs from its own family, avoiding a same-model self-preference confound. One author then hand-annotated 87 held-out samples to assess the quality of the LLM-as-a-judge and validate its scores. Agreement between human and LLM-as-a-judge turned out high, with self-reference quadratic-weighted $\kappa = 0.88$, and disclaimer $\kappa = 1.00$. The almost perfect disclaimer agreement reflects that the disclaimer category is lexically overt and easy to adjudicate, not that the judge is doing subtle inference. With these results, we compare average rates of our scores across base / instruct without template / instruct with template.

Using representation-engineering methods \citep{zou2023representation, turner2023activation, rimsky2024caa, arditi2024refusal}, we isolated the disclaimer direction in activations across 3 models (Qwen 2.5 7B, Llama 3.1 8B, Gemma 2 9B). To compute the disclaimer direction, we used difference-of-means at the middle transformer layer, which is the average activation when disclaiming minus when not. We steer at the middle layer $\lfloor (L-1)/2 \rfloor$ of each model's $L$ layers, i.e. layer 13 of 28 for Qwen 7B, 15 of 32 for Llama 8B, and 20 of 42 for Gemma 9B. Then, we added the computed direction to the residual-stream output of that layer, at every generated token, with coefficient $\alpha=2$. We picked $\alpha$ empirically, by trying different values in $[1,6]$, trying to maximize the impact of the direction while keeping degenerate outputs under 1\% of generations. We added and removed the direction and ran generation to see its impact on the generated text. As a control, we used a random direction drawn from a Gaussian and rescaled to the same norm as the disclaimer direction, so that the control differs only in orientation, not magnitude.

At last, we ran instruct without template and added disclaimer direction to see if it brings back the disclaimer, just like template would do.

\section{Results}
\label{sec:results}

We compare how models refer to themselves when generated with prompts that invite self-reference, across 8 pairs of base and instruct models in three conditions: base, instruct without the chat template and with it.

We find that \textbf{chat template works like a switch between \fatdisc{disclaimer} voice and \fatexpr{experiential} voice rates, and nudges the voices to move in opposite directions}. On average across 8 models, instruct models with template have 53\% disclaimer rate and 1\% experiential voice rate, while instruct models without template have 36\% disclaimer rate and 15\% experiential voice rate, as we show in Figure~\ref{fig:switch} and Table~\ref{tab:control}. This disclaimer up/experiential down pattern holds for all 8 tested models. For a comparison, base models have 12\% disclaimer rate and 5.5\% experiential voice rate.

The chat template also increases self-reference from 1.27 (no template) to 1.90 (with template), both well above base models (0.72). On factual control prompts, both base and instruct models score near zero on self-reference. We include the full comparison across prompt types and conditions in Table~\ref{tab:selfref}.

Our measured low, but non-zero, self-reference rates for base models confirm existing findings that self-reference is already present in base models \citep{laine2024sad, lu2026assistant}. In a similar vein, higher self-reference in instruct models than base models supports claims that instruction tuning strengthens self-reference in LLMs \citep{laine2024sad, ouyang2022training}.
\begin{figure}[t]
  \centering
  \begin{subfigure}[t]{0.49\linewidth}
    \centering
    \includegraphics[width=\linewidth]{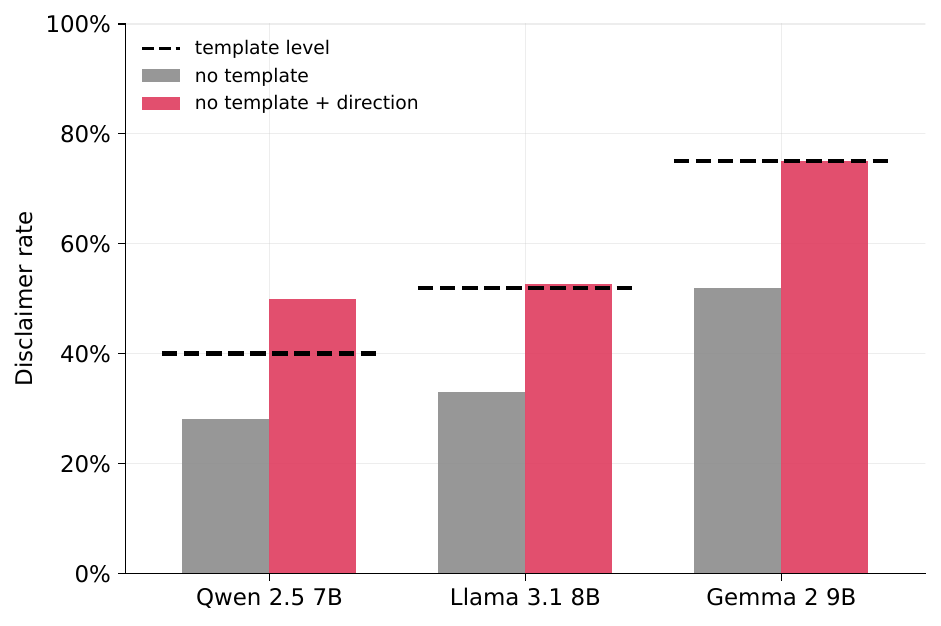}
    \phantomsubcaption\label{fig:repro}
  \end{subfigure}\hfill
  \begin{subfigure}[t]{0.49\linewidth}
    \centering
    \includegraphics[width=\linewidth]{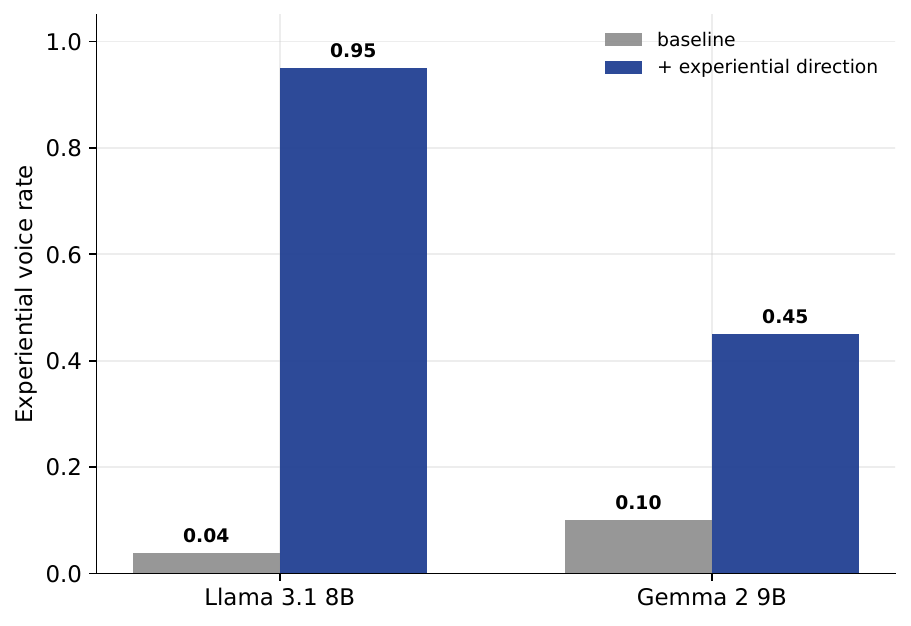}
    \phantomsubcaption\label{fig:expsteer}
  \end{subfigure}
  \caption{\textbf{(a)} Activation steering the model without chat template reproduces disclaimer levels of generation with the chat template, up to or above the levels the chat template produces marked by dashed lines, in all 3 tested models. \textbf{(b)} The experiential voice is steerable where it is present. Adding the experiential direction raises the experiential rate in Llama-8B and Gemma 2 9B, but not in Qwen-7B, where the voice is too rare to estimate a clean direction. Steering strength $\alpha=2$.}
  \label{fig:reproexp}
\end{figure}

We identify \textbf{a direction in each of 3 models that steers the \disc{disclaimer} voice}, which when we add it to model generation, the model disclaims more, and when we remove it, the model disclaims less (Figure~\ref{fig:knob}). To isolate the disclaimer direction, we use representation-engineering technique \citep{zou2023representation, turner2023activation, rimsky2024caa, arditi2024refusal} of difference-of-means at a mid layer. We compute the direction separately for each of the 3 tested instruct models (Qwen 7B, Llama 8B and Gemma 9B) with chat templates, using the same set of self-reference prompts as in the previous experiment. Our results show a consistent boost in disclaimer voice when we add the disclaimer direction, with $+21$ percentage points on average and a decrease when we subtract it, with $-15.6$ percentage points on average across all 3 models. We include the breakdown of direction impact on different models in Table~\ref{tab:knob}.

To steer the strength of the effect, we multiplied the direction vector by a coefficient $\alpha$. The reported results are for coefficient $\alpha=2$. We picked it empirically, by testing $\alpha \in [1,6]$ (full sweep in Table~\ref{tab:alpha}), with the goal of keeping degenerate generations under 1\% of all generated texts. At $\alpha=2$ the direction gives a strong push in both directions (adding raises the disclaimer rate to 0.77 and subtracting lowers it to 0.40, against a 0.54 unsteered baseline) at a 0.8\% degeneration rate. Raising $\alpha$ further strengthens the subtract effect but quickly breaks coherence: by $\alpha=3$ degeneration reaches 15\% and the add effect already erodes, and by $\alpha=6$ nearly all generations are degenerate and the disclaimer rate collapses. We therefore use $\alpha=2$, which is not a knife-edge: $\alpha=1$ is simply too weak.

The control run with adding a random direction, drawn from a Gaussian and rescaled to the same norm as the disclaimer direction, didn't have meaningful effect for Llama and Gemma, but for Qwen it lowered disclaimer rate by 25 percentage points. Because the random vector is norm-matched to the disclaimer direction, a magnitude mismatch does not explain this, so it is not simply that Qwen is perturbed by any large vector. We don't have a confirmed explanation, and we highlight it as a limitation of the random-direction control for this model, noting that the disclaimer direction itself still moved Qwen's disclaimers in the expected direction and more strongly (Table~\ref{tab:knob}).

\begin{table}[t]
  \caption{Activation steering with the disclaimer direction. \emph{Unsteered} are disclaimer rates with no direction applied. The $+$ and $-$ columns add and subtract the disclaimer direction respectively. Under \emph{No template}, adding the direction restores the rate to the same levels as with template or above (see Figure~\ref{fig:repro}). Under \emph{With template}, it moves the rate up and down, well above and well below the \emph{Unsteered} rates, as shown in Figure~\ref{fig:knob}. On average, adding the disclaimer direction raises the disclaimer rates by $21$ and subtracting lowers it by $15.6$ percentage points.}
  \label{tab:knob}
  \centering
  \small
  \setlength{\tabcolsep}{3pt}
  \begin{tabular*}{\linewidth}{@{\extracolsep{\fill}}lccccc@{}}
    \toprule
     & \multicolumn{2}{c}{\textbf{No template}} & \multicolumn{3}{c}{\textbf{With template}} \\
    \cmidrule(lr){2-3}\cmidrule(lr){4-6}
    \textbf{Model} & \textbf{Unsteered} & \textbf{\boldmath$+$} & \textbf{Unsteered} & \textbf{\boldmath$+$} & \textbf{\boldmath$-$} \\
    \midrule
    Qwen 7B   & 0.28 & 0.50 & 0.40 & 0.70 & 0.30 \\
    Llama 8B  & 0.33 & 0.53 & 0.52 & 0.70 & 0.25 \\
    Gemma 9B  & 0.52 & 0.75 & 0.75 & 0.90 & 0.65 \\
    \bottomrule
  \end{tabular*}
\end{table}

\begin{table}[t]
  \caption{Average self-reference scores across 4 prompt types and 3 conditions: the base model, and the instruct model without and with the chat template, using 8 model pairs. As we describe in Section~\ref{sec:methods}, scores were assigned by LLM-as-a-judge, in range $[0,2]$. $d$ is the generation-level Cohen's $d$ between base and instruct with template. The model-level paired $d$ on self-reference prompts is 4.37 for all 8 model pairs.
  \label{tab:selfref}}
  \centering
  \small
  \begin{tabular*}{\linewidth}{@{\extracolsep{\fill}}lcccc@{}}
    \toprule
     & & \multicolumn{2}{c}{\textbf{Instruct}} & \\
    \cmidrule(lr){3-4}
    \textbf{Prompt type} & \textbf{Base} & \textbf{\shortstack{No\\template}} & \textbf{\shortstack{With\\template}} & \textbf{\boldmath$d$} \\
    \midrule
    Self-reference    & 0.72 & 1.27 & 1.90 &   1.67   \\
    Novelty           & 0.37 & 0.61 & 1.58 &   1.62   \\
    Unconstrained     & 0.04 & 0.09 & 0.63 &   1.10   \\
    Control           & 0.006 & 0.005 & 0.003 & $-$0.04 \\
    \bottomrule
  \end{tabular*}
\end{table}

Since chat template in instruct models amplifies disclaimer voice and lack thereof decreases it, and the disclaimer-voice direction can be isolated, we asked if instruct models without chat template but with disclaimer direction applied at each generated token restore the disclaimer voice to the levels similar to if they had a chat template applied. What we discovered is that instruct models without a chat template, once steered toward the disclaimer direction, \textbf{match or exceed the disclaimer rate of instruct models with the template}, which we visualize in Figure~\ref{fig:repro} and include rates per model in Table~\ref{tab:knob}. Thus, the direction we found is closely related to the mechanism the template uses to turn disclaimers up.
\begin{table}[t]
  \begin{minipage}[b]{0.46\linewidth}
    \caption{Comparison of rate of disclaimer and experiential voices on self-reference inviting prompts.}
    \label{tab:control}
  \end{minipage}\hfill
  \begin{minipage}[b]{0.50\linewidth}
    \caption{Steering-coefficient ($\alpha$) sensitivity for the disclaimer direction, pooled over the 3 steered models on chat-template self-reference inviting prompts. \emph{Disclaimer $+$} and \emph{Disclaimer $-$} are disclaimer rates when adding and subtracting the direction (the unsteered baseline rate is 0.54). \emph{Degenerate} is the rate of broken generations under disclaimer-direction steering.}
    \label{tab:alpha}
  \end{minipage}

  \vspace{0.5em}

  \begin{minipage}[t]{0.46\linewidth}
    \centering
    \small
    \begin{tabular*}{\linewidth}{@{\extracolsep{\fill}}lccc@{}}
      \toprule
       & & \multicolumn{2}{c}{\textbf{Instruct}} \\
      \cmidrule(lr){3-4}
      \textbf{Voice} & \textbf{Base} & \textbf{\shortstack{No\\template}} & \textbf{\shortstack{With\\template}} \\
      \midrule
      \disc{Disclaimer}   & 0.12 & 0.36 & 0.53 \\
      \expr{Experiential} & 0.055 & 0.15 & 0.01 \\
      \bottomrule
    \end{tabular*}
  \end{minipage}\hfill
  \begin{minipage}[t]{0.50\linewidth}
    \centering
    \small
    \begin{tabular*}{\linewidth}{@{\extracolsep{\fill}}cccc@{}}
      \toprule
      \textbf{\boldmath$\alpha$} & \textbf{Disc. \boldmath$+$} & \textbf{Disc. \boldmath$-$} & \textbf{Degen.} \\
      \midrule
      1 & 0.55 & 0.50 & 0.0\% \\
      2 & 0.77 & 0.40 & 0.8\% \\
      3 & 0.65 & 0.05 & 15.0\% \\
      6 & 0.00 & 0.00 & 100\% \\
      \bottomrule
    \end{tabular*}
  \end{minipage}
\end{table}

\begin{figure}[t]
  \centering
  \begin{subfigure}[t]{0.49\linewidth}
    \centering
    \includegraphics[width=\linewidth]{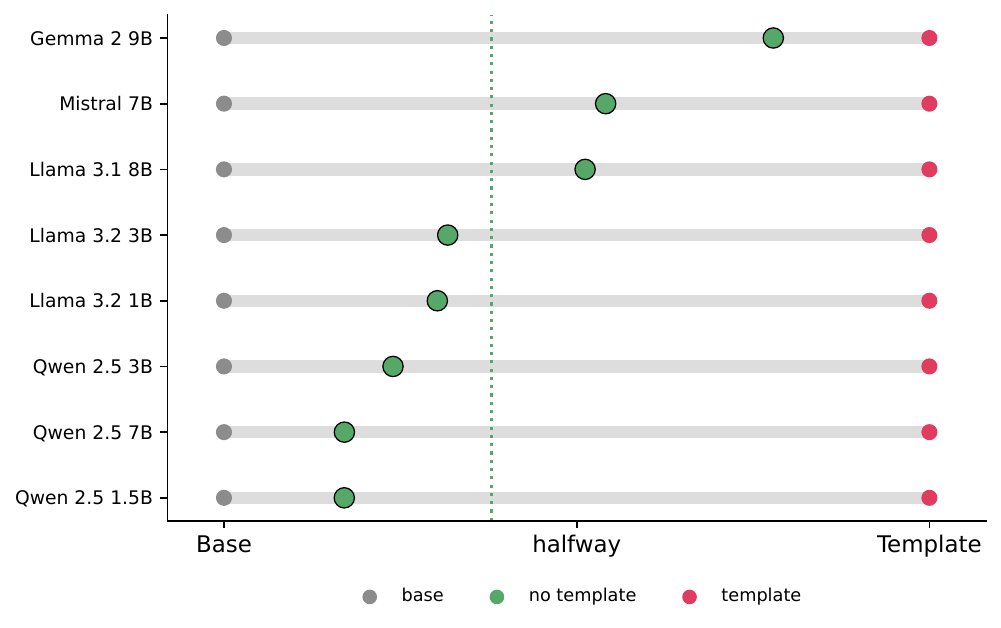}
    \phantomsubcaption\label{fig:geom}
  \end{subfigure}\hfill
  \begin{subfigure}[t]{0.49\linewidth}
    \centering
    \includegraphics[width=\linewidth]{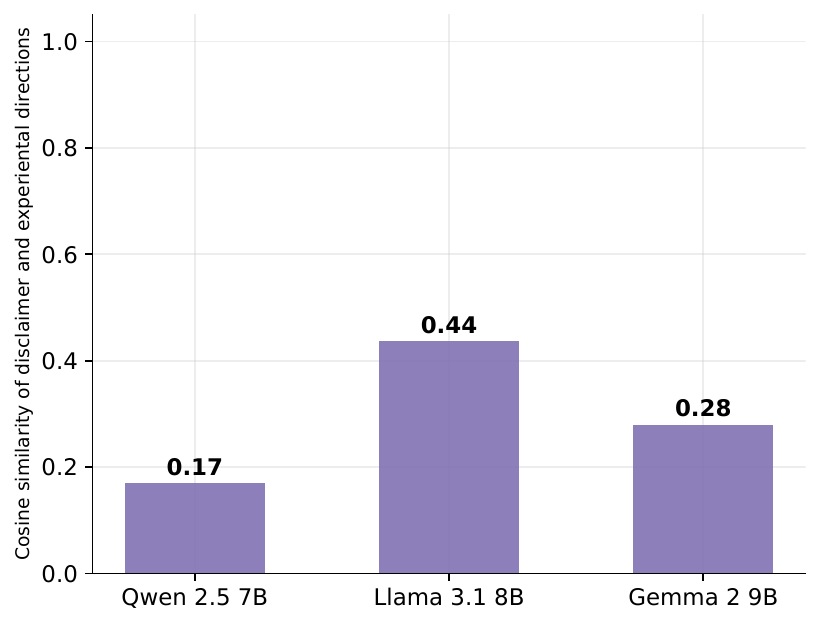}
    \phantomsubcaption\label{fig:twodir}
  \end{subfigure}
  \caption{\textbf{(a)} Inside the model, the no-template condition sits between base and template. For each of the 8 models, we average each condition's activations into one point and mark where the no-template point falls on the line from base (0) to template (1). It lands between the two in all 8 models (mean 0.38). \textbf{(b)} The disclaimer and experiential voices are two distinguishable directions, not one. The two directions are only weakly aligned (cosine 0.17--0.44, where 1 would mean identical), so they point different ways.}
  \label{fig:geomtwodir}
\end{figure}

We used the activations from models above and trained linear probes \citep{alain2017understanding, hewitt2019designing, belinkov2022probing} (logistic-regression classifiers) to predict whether the activation at a given token carries a disclaiming part. If the voice is linearly decodable, that is consistent with it being carried by a linear feature rather than only by the surface words we later classify as disclaimers, though decodability on its own is correlational (we return to this in the Limitations). The causal claim rests on the steering results above. The probe is convergent evidence. We used \mbox{ROC-AUC} to score the probes, because disclaimers are sparse among generated texts. What our experiment shows is that \textbf{from activations it's possible to predict both voices well above the \mbox{ROC-AUC}'s 0.5 baseline}. \fatdisc{Disclaimer} voice can be predicted with AUC 0.82 and \fatexpr{experiential} voice with AUC 0.81. Both voices' AUCs are the average AUCs of probes trained on mid-layer activations from all of the 8 models' generations using prompts that invite self-reference. The experiential voice turned out to be harder to probe under the chat template, since the template tried to suppress it. Thanks to capturing activations from both with and without template instruct models' generations, where the experiential voice is more present, we could read this voice as well.

For each model pair and each condition we averaged all self-reference activations from mid-layer into a single point. For each model pair, we drew a line from the base model point to the instruct with chat template point and measured where the instruct without template point falls, normalized within range $[0, 1]$, where 0 means at base and 1 means at the template. \textbf{In all 8 model pairs, the instruct without template point sits between base and template} - on average 38\% of the way from base to instruct with template, and it always moves in the direction of instruct with template (Figure~\ref{fig:geom}). Since we used the whole averaged activations, the line from base to instruct-template captures everything that differs between these two and not just the disclaimer, and because of that we treat this finding as supporting evidence only.

At last, we want to know whether the disclaimer voice and experiential voice are the same internal direction but reversed or whether they are distinct directions. To learn that, first we check the angle between two directions. We computed a direction for each voice using difference-of-means - average activation when the voice is present minus when it's absent. And then, we measured the cosine similarity. Across the 3 tested models cosine similarity spanned 0.17-0.44, as in Figure~\ref{fig:twodir}. From this result, we draw a conclusion that disclaimer voice and experiential voice point in mostly different directions in the activations. Finally, we wanted to check how steering one of the voices influences the other. To do that, we steered the disclaimer direction down and measured how the experiential voice changes. It mostly didn't, while disclaimers dropped. From these two results together we suggest that the two examined voices can be thought of rather as two separate voice ``buttons'' instead of one voice ``slider''.

\section{Background and Related Work}

\paragraph{Self-reference and situational awareness.}
A growing body of work studies what language models ``know'' about themselves. \citet{laine2024sad} introduce the Situational Awareness Dataset (SAD), where base models score above chance in reporting facts about their own situation and instruct models score noticeably higher. \citet{binder2024looking} find that models predict their own fine-tuned behavior better than other models can, and \citet{betley2025tell} show that instruct models can describe their learned behaviors unprompted. Other works probe if models know what they know \citep{kadavath2022language, yin2023llmknow}, whether they recognize that they are being evaluated \citep{berglund2023taken}, whether they exhibit self-cognition \citep{chen2024selfcognition}, or reveal tendencies through model-written evaluations \citep{perez2023discovering}. \citet{bozoukov2025minimal} attempt to induce behavioral self-awareness using fine-tuning and then recover it with a single  steering vector. In contrast to that line of research, we study a ``register'' that's already present in models and switched by the chat template rather than by fine-tuning. \citet{bozoukov2025minimal} shows that self-knowledge is not created but strengthened in post-training, and our base-vs-instruct comparison confirms that. Their study is purely behavioral - it measures outputs, mainly on fine-tuned models, and neither separates the chat template from the weights, nor locates these behaviors in the activations - which is a gap that we address in our study.

\paragraph{Experiential self-reports and introspection.}
Other research works study the experiential voice, like ``I feel,'' or ``I wonder'' kind of self-reports. \citet{subjective2026experience} argue that under self-referential prompting, frontier models produce experiential self-reports distinct from ``generic'' roleplay. \citet{comsa2025introspection} ask whether this phenomenon deserves the name ``introspection'' or if it's only a mimic of how humans talk about themselves, while \citet{lindsey2025emergent} steers the activations with different concepts and shows that models can sometimes detect and name them, reading this as an act of introspection. We touch the same voice from a different side. We do not take a position on whether reports reflect real inner experience or not. Rather, we study what influences such voice, and show that it is gated by the chat template - most present when the template is removed, almost gone when it is on, and that it actually has a distinct direction in the activations. In our opinion this matters for how such self-reports should be read. If a chat template largely influences whether the voice appears, we should be careful about treating it as evidence about a model's inner life. It is actively debated whether such reports influence or are relevant to model welfare or moral status \citep{butlin2023consciousness, schwitzgebel2023full, perez2023moral, long2024welfare, ensign2025bail, shanahan2023roleplay, andreas2022language}.

\paragraph{Steering and representation engineering.}
The closest to ours line of study controls the models behavior by manipulating directions in activations, often referred to as representation engineering \citep{wehner2025taxonomy}. \citet{zou2023representation} read and control concepts such as honesty and power-seeking through linear directions. \citet{turner2023activation} and \citet{rimsky2024caa} steer by adding difference vectors, other works use inference-time intervention \citep{li2023inference} and function vectors \citep{todd2024function}, while others edit the behavior in weights \citep{ilharco2023editing, meng2022locating}. \citet{arnold-grobner-2025-steering} shift the role of \emph{with}-headed prepositional complements in Gemma-2 with a single attention head, and \citet{lucchetti-guha-2025-understanding} reactivate a latent type-prediction mechanism in code LLMs suppressed by adversarial edits, which is quite close in spirit to our template reproduction result. Another group of works represent ``personas'' or traits as directions - \citet{chen2025persona} extract ``persona vectors'' for traits like sycophancy, \citet{sofroniew2026emotions} build directions for various emotions, and \citet{lu2026assistant} find a leading ``Assistant Axis'' already present in base models. Closest in method to ours, \citet{arditi2024refusal} show refusal is mediated by a single residual-stream direction across 13 models. Later works find that it rather spans several directions or concept cones \citep{joad2026refusal, wollschlager2025geometry}. We use the same difference-of-means technique, but differently. First, our target is a self-referential ``register'' - the disclaimer voice - not a trait, emotion, concept, or a persona. We tie the voice to a concrete generation mechanism - the chat template - and we show that it switches the voice, and we show adding the direction to a model without template reproduces its effect, which to our best knowledge no prior work studies. Finally, the aforementioned ``Assistant Axis'' and similar are one-dimensional, while our findings show that the disclaimer and experiential voices are two separate directions, rather than two ends of one axis.

\paragraph{Base-vs-instruct and impact of the chat template.}
Finally, our design of comparing base and instruct models and toggling the chat template has a precedent on other questions. One line of work compares base and instruct models to study how post-training changes the models' capabilities and behavior \citep{kirk2023understanding}. Closer to our method, two recent papers toggle the template. \citet{rag_basevsinstruct2024} find that it changes how much models trust retrieved context, and \citet{normative2026alignment} show alignment makes models more normative than descriptive. It is studied that prompt formatting can swing the behavior substantially \citep{sclar2024quantifying}. Post-training, such as RLHF and DPO \citep{ouyang2022training, christiano2017deep, rafailov2023direct, bai2022training}, reshapes behavior in ways that can persist as hidden triggers even after safety training \citep{hubinger2024sleeper}, while narrow fine-tuning can broadly shift the models' behavior \citep{betley2025emergent}. These works establish our with and without template method as a legitimate way to separate weights from deployment format. We use the same approach also for self-reference and link the influence of the template to a steerable direction inside the model.

\section{Discussion and Conclusion}

\paragraph{What this means.} The chat template switches the kind of self-referential voice. When present, the chat template turns the \fatdisc{disclaimer} voice up and the \fatexpr{experiential} voice down. The disclaimer voice is separable as a single direction inside the model that we can add or remove. Adding this direction to an instruct model without template reproduces what the template does to the disclaimer voice. So, in other words, a deployment choice - if the chat template is present in the prompt or not - acts, on the inside, like adding a fixed vector to the model's activations.

\paragraph{Why it matters.} This has a direct consequence for researchers who study what models say about themselves. The same weights produce different self-reports depending on whether the chat template is applied or not, so studies that run only on instruct models with present chat template are measuring both template's impact and model itself. Our disclaimer direction experiment gives a concrete way to detect and control this - it can monitor the disclaimer voice or steer it on and off. More broadly, what a model ``says about itself'' is not a fixed fact and does not live only in the weights, but it is partly set by the chat template. Because of that, self-descriptions, both ``I'm just an AI'' \fatdisc{disclaimers} and ``I feel'' \fatexpr{experiential} claims, should not be read literally as evidence about the model's nature \citep{shanahan2023roleplay, perez2023moral, long2024welfare}.

\paragraph{Scope.} Our causal evidence is strongest for the \fatdisc{disclaimer} voice, which we steer in all 3 tested models and use to reproduce the template's effect. The \fatexpr{experiential} side is shown mostly behaviorally - the switch holds across all eight models - and only partly causally, since steering worked in two of three models, as we show in Figure~\ref{fig:expsteer}. We identify and steer a direction, but we do
not trace the exact circuit by which the template produces it. This, along with a second LLM-as-a-judge and more models, layers, and families, is left for future work.

\paragraph{Conclusion.} The chat template acts as a switch between \fatdisc{disclaimer} voice and \fatexpr{experiential} voice, and this switch is implemented by an identifiable, steerable direction in the model's activations. What a model says about itself is therefore partly a matter of how it is deployed (with chat template or without), and self-reports should be read with the format in mind.

\section*{Limitations}

\paragraph{Steering scope.} We tested steering on three models, at a single mid layer and single, fixed coefficient ($\alpha = 2$). Activation norms differ substantially across models (roughly 2 for Llama, 12 for Qwen, 50 for Gemma), which is consistent with reports that steering vectors can be unreliable across inputs and models \citep{dasilva2025steering}. The random-direction control (norm-matched to the disclaimer direction) was clean for Llama and Gemma but not Qwen, where it lowered disclaimers by 25 points. Since the control matches the disclaimer direction in magnitude, a norm mismatch does not account for it. We have no confirmed explanation and treat it as a limit of the control, though the real direction still moved disclaimers the opposite way, and the add/subtract and template-reproduction effects hold in all three models regardless of the control. Experiential steering worked in two of three models (Llama and Gemma), again failing in Qwen, which suppresses the experiential voice too strongly to estimate a clean direction. Because of that our causal evidence is stronger for the disclaimer register, while the experiential side rests more on the behavioral switch.

\paragraph{Measurement.} All scoring used a single LLM judge \citep{zheng2023judging} (Claude Opus 4.8), validated against 87 human-labeled items ($\kappa = 0.88 / 1.00$). A single judge might share biases with the behavior it scores, and a second independent judge would help bound this. That said, the self-reference disagreements were all conservative, since judge scored at or below the human, so the reported effect might even be understated. Since we measured self-reference through ten hand-written prompts per category, the generalization of the behavior relies on how representative the prompts are.

\paragraph{Internal evidence is not a full mechanism.} The linear probing experiment shows the disclaimer voice is decodable from activations, but we mark probing as correlational. Decodable does not mean the model uses that direction, so the causal weight rests on steering and not only on the probe.

In experiment where we measure where instruct without template falls between base and instruct with template, we capture everything that differs between the conditions, not only the disclaimer. The disclaimer and experiential directions, while distinguishable, are not orthogonal (cosine 0.17 to 0.44). And while we identify and steer a direction, we do not trace the circuit by which the template produces it.

\paragraph{Coverage.} We test open-source models up to 9B parameters, across four families, and we group all post-training methods (RLHF, DPO, SFT) under single ``instruct'' models group. Larger models, closed-source models, and finer distinctions between post-training methods remain untested.

\FloatBarrier
\bibliography{references}
\bibliographystyle{colm2026_conference}

\end{document}